# On the Efficacy of Metrics to Describe Adversarial Attacks

Tommaso Puccetti, Tommaso Zoppi, Andrea Ceccarelli

Department of Mathematics and Informatics, University of Florence, Italy

**Abstract**

Adversarial defenses are naturally evaluated on their ability to tolerate adversarial attacks. To test defenses, diverse adversarial attacks are crafted, that are usually described in terms of their evading capability and the $L_0$, $L_1$, $L_2$, and $L_\infty$ norms. We question if the evading capability and L-norms are the most effective information to claim that defenses have been tested against a representative attack set. To this extent, we select image quality metrics from the state of the art and search correlations between image perturbation and detectability. We observe that computing L-norms alone is rarely the preferable solution. We observe a strong correlation between the identified metrics computed on an adversarial image and the output of a detector on such an image, to the extent that they can predict the response of a detector with approximately 0.94 accuracy. Further, we observe that metrics can classify attacks based on similar perturbations and similar detectability. This suggests a possible review of the approach to evaluate detectors, where additional metrics are included to assure that a representative attack dataset is selected.

## Introduction

DNNs are vulnerable to adversarial attacks, in which the input (e.g., images, texts, tabular data) is deliberately modified to mislead the target model (Carlini et al. 2019), (Goodfellow et al. 2014). These attacks are particularly effective in deceiving an image classifier to produce completely wrong predictions.

Consequently, different defenses have been proposed in the last years. Typically, defenses are evaluated based on their ability to protect a target model against adversarial attacks. Attacks are crafted for the target model, starting from the input dataset, and consist of multiple adversarial images. Each adversarial image is usually described using i) the attack configuration parameters, ii) the success rate, and iii) values of the $L_0$, $L_1$, $L_2$, and $L_\infty$ norms (simply named L-norms in what follows). L-norms are the typical means to describe the perturbation introduced in the input by the application of an attack (Carlini et al. 2019), (Xu et al. 2018), (Meng et al, 2017).

Information on the attacks is particularly important to evaluate the quality of a defense. The attacks used to evaluate defenses should be sufficiently representative of the possible attacks and the possible perturbations introduced in the image. This should be evident from the description of the evaluation approach applied: otherwise results on the robustness of the defenses could be not conclusive, specific to the considered attack sets, or difficult to interpret.

We investigate whether computing the perturbation through different metrics is a useful proxy to use in the evaluation of adversarial defenses.

Intuitively, the sole usage of L-norms may be misleading because a defense may detect attacks of type A and fail to detect attacks of type B, even if they have similar L-norms values. If only type A attacks are used, the defensive mechanism may be erroneously considered very robust whereas it cannot reliably detect type B attacks.

In this paper, we compute L-norms and other metrics from image quality through an extensive experimental campaign. The objective is to understand whether the effectiveness of an attack against a defensive solution can be established by quantifying the perturbation introduced in an image. As such, we compute and analyze metrics to identify those that identify a strong correlation between the perturbation and the prediction outcome of two state-of-the-art defenses.

Our empirical assessment is structured as follows. We generate 10 attack sets from 10 different attacks. For each adversarial image, we compute a total of 12 metrics: 4 L-norms and 8 image quality metrics from the literature, namely MSE, UQI, ERGAS, SAM, SCC, VIFP, RASE, and PSNR-B. We train and exercise two state-of-the-art detectors to detect attacks on each attack set. We train a Random Forest (RF) classifier using the values of the 12 metrics as features and the output of the detector as ground truth. As such, the RF classifier provides a way to correlate the perturbation introduced in the image to the ability of the attack to deceive defenses. Our analysis highlights that using image quality metrics as features allows the classifier to predict the output of a detector with very high accuracy, even better than L-norms. Further, training the same RF classifier with a leave-one-out approach shows that image quality metrics and L-norms can identify classes of attacks that introduce similar perturbations and lead to similar detection results. This is a stepping stone towards partitioning adversarial attacks into classes, from which representatives should be selected when testing defenses.

# Background

## Adversarial Attacks on Image Classifiers

We focus only on evasion adversarial attacks that target image classifiers. This is the typical application domain in research on adversarial attacks, and it is the focus of most of the works on the subject (Carlini et al. 2019), (Goodfellow et al. 2014), (Madry et al. 2017). An evasion attack aims at perturbating an image x, correctly classified, in such a way that the resulting image x' is wrongly classified, while being perceptually indistinguishable from the original.

An image classifier model can be seen as a function: $f(x, \theta): R^{h \times w \times c} \rightarrow \{1 \ldots k\}$ that maps an input image x to a finite label set Y with k classes. The $\theta$ indicates the parameters of the model f, and h, w, and c represent image height, width, and channel, respectively. Usually, the adversary generation method consists in searching for a perturbation $\delta \in R^{h \times w \times c}$ that maximizes the loss function L. Therefore, $\delta$ is estimated as $\delta^*$: argmax ($L_p(\theta, x+\delta, y)$), where y is the label of x, and p can be 0, 1, 2 and $\infty$. The adversarial counterpart of x is expressed as x': $= x + \delta^*$.

Evasion attacks can be white-box or black-box. White-box attacks imply that the attackers have full knowledge of the target model, while black-box attacks do not have this information.

## Metrics to select adversarial attacks

Protections against evasion attacks are naturally evaluated in terms of their ability to detect or tolerate attacks. The attacks are typically crafted according to selected methodologies and tools. In addition, configuration parameters of the attacks are needed for reproducibility. The configuration parameters can change according to the attack method and the target classifier. For example, the FGM (Goodfellow et al. 2014), and BIM (Kurakin, 2017) attacks require the specification of the maximum number of iterations and the maximum amount of perturbation epsilon allowed. Lastly, the perturbation introduced by the attack can be measured to have precise information about how far the adversarial image is from the original one. The works in this field usually report on i) configuration parameters used to run the attack method, ii) the perturbation introduced by the attacker, and iii) the attack success rate (Carlini et al. 2019), (Xu et al. 2018), (Meng et al, 2017). This way, comparing defensive solutions is possible.

At the state-of-the-art, the amount of perturbation introduced by an attacker is quantified using the pixel $L_p$ norms, for any p > 0:

$$\|v\|_p = \left( \sum_{i=1}^{N} |v_i|^p \right)^{\frac{1}{p}}$$

where v=∥x–x'∥ is the perturbation introduced in an image x to obtain his adversarial counterpart x', and $v_i$=∥$x_i$–$x_i$'∥ is the difference pixel by pixel of the two images, with n the total number of pixels. It is common practice to report the value of one of the $L_\infty$, $L_0$, $L_1$, and $L_2$ to describe the composition of the attack set. Each of these L-norms describes different aspects of the perturbation through a single value, thus, are informative in different ways, as reported by (Carlini et al. 2019):

- $L_0$ distance measures the number of coordinates i such that $x_i \neq x_i'$. The $L_0$ distance is equal to the number of pixels that have been altered in an image.
- $L_1$ distance, also known as Manhattan Distance, is defined as the sum of the absolute difference between pixels of two images.
- $L_2$ distance, also known as Euclidean distance, is defined as the squared root of the sum of the squared absolute difference between pixels of two images. The $L_2$ distance can remain small when there are many small changes to many pixels.
- $L_\infty$ distance is defined as the largest absolute difference between pixels of two images. It measures the maximum change to any of the coordinates: $\|x - x'\|_\infty = \max(|x_i - x_i'|, \ldots, |x_n - x_n'|)$. For images, we can think there is a maximum budget, and each pixel is allowed to be changed by up to this limit, with no limit on the number of pixels that are modified.

## Image Quality Metrics

We identify alternative metrics to quantify the perturbation with the aim of extrapolating as much information as possible from the adversarial image. We explore the image quality domain because particularly fits our necessity. Image quality metrics have been crafted to quantitatively evaluate the perceived quality of images modified by a variety of distortions (e.g., processing, compression, etc.). This is comparable to introducing an adversarial perturbation into an image. State-of-the-art image quality metrics can be classified according to the availability of an original (distortion-free) image to use as a comparison. In this paper, we consider only the case when the original image is available: this is called the *full reference approach*. The rationale is that we want to use the original image as a reference to compute the adversarial perturbation. We selected the following metrics.

*Mean Squared Error (MSE)* (Yim et al. 2010) is usually used to compute an estimator's quality. In our context, the vector $x_i$ and $x_i'$ can be seen as a linearized version of original and adversarial images.

$$\text{MSE} = \frac{1}{n} \sum_{i=1}^{n} (x_i - x_i')$$

*Universal Quality Image Index (UQI)* (Wang et al. 2002) calculates the amount of transformation of relevant data from the reference image into the perturbed image. The range of this metric is -1 to 1. The value 1 indicates that the reference and perturbed images are similar.

$$UQI = \frac{4\sigma_{xx'}\bar{x}\bar{x}'}{(\sigma_x^2 + \sigma_{x'}^2)[(\bar{x})^2 + (\bar{x}')^2]}$$

*Erreur Relative Globale Adimensionnelle Synthèse (ERGAS)* (Wald et al. 2000) calculates the average error of each band of the perturbed image with respect to the reference one. High values of ERGAS indicate the low quality of the perturbed image, while lower values indicate good quality.

$$ERGAS = 100 \frac{h}{w} \left[ \frac{1}{n} \sum_{k=1}^{n} \left( \frac{RMSE^2}{mean^2} \right) \right] \frac{1}{n}$$

where h and l are height and width of the image and the RMSE is the Root Mean Squared Error computed between original and altered image.

*Spectral Angle Mapper (SAM)* (Yuhas et al. 1992) computes the spectral angle between the pixel vector of the reference image, and of the perturbed image. It is worked out in either degrees or radians. It is performed on a pixel-by-pixel basis. A value of SAM equal to zero denotes the absence of spectral distortion.

$$SAM = arccos\left(\frac{\langle x, x' \rangle}{||x||_2 \cdot ||x'||_2}\right)$$

*Spatial Correlation Coefficient (SCC)* (Zhou et al. 1998) represents the correlation between two visual signals of images in a cortical visual space.

$$SCC = \frac{\sum_{m-1}^{M}\sum_{n-1}^{N}(x_{\min} - \bar{x})(x'_{\min} - \bar{x}')}{\sqrt{\left(\sum_{m-1}^{M}\sum_{n-1}^{N}(x_{\min} - \bar{x})^2\right)\left(\sum_{m-1}^{M}\sum_{n-1}^{N}(x'_{\min} - \overline{x'})^2\right)}}$$

*Relative Average Spectral Error (RASE)* (Gonzalez et al. 2004) determines the difference in spectral information between each band of the merged image and of the original image. Given M the mean radiance of the N spectral bands $B_i$ of the original image, and the RMSE the root mean square error computed in following the expression:

$$RASE = \frac{1}{M}\sqrt{\frac{1}{N}\sum_{i=1}^{N}RMSE^2(B_i)}$$

*Visual Information Fidelity (VIF)* (Sheik et al. 2011) is computed for a collection of wavelet coefficients from each sub-band that could either represent an entire sub-band of an image, or a spatially localized region of sub-band coefficients. In the former case, the VIF quantifies the information fidelity for the entire image.

$$VIF = \frac{\sum_{j \in subbands} I(\vec{C}^{N,j}; \vec{F}^{N,j} \mid s^{N,j})}{\sum_{j \in subbands} I(\vec{C}^{N,j}; \vec{E}^{N,j} \mid s^{N,j})}$$

*Block Sensitive - Peak Signal-to-Noise Ratio (PSNR-B)* (Yim et al. 2010) is a widely used metric. It is computed by the number of gray levels in the image divided by the corresponding pixels in the original and the perturbed images. When the value is high, the perturbed and original images are similar.

$$PSNR\text{-}B(x, x') = 10 \log_{10} \frac{L^2}{MSE - B(x,x')}$$

Where $L^2$ is the square of the numbers of pixels in the image

## Methodology and Research Questions

### Selection of Dataset and Classifier

We select the CIFAR-10 (Krizhevsky et al. 2009) dataset which is well-known at the state of the art, color based, and images are of limited size. The size of the images is 32×32×3 and each image uses a 24-bit representation of each pixel. Then we select the ConvNet12 CNN from (Ma et al. 2018) as target classifier because it is fast and of reasonable accuracy (0.85). The model has 6 convolutional layers, 1 dense layer, 3 pooling layers, and 3 dropout layers. The trainable parameters are 2,923,050.

### Generation of Adversarial Image Sets

As suggested in (Carlini et al. 2019), we consider white box and black box attacks to build candidate attack sets. We select 7 white-box attacks and 3 black-box attacks among those implemented by the Adversarial Robustness Toolbox ART (Nicolae et al. 2019). More specifically, we use the white-box attacks: CW2-Carlini and Wagner L2 (Carlini et al. 2019), FGM-Fast Gradient Method (Goodfellow et al. 2014), BIM-Basic Iterative Methods (Kurakin et al. 2018), DEEP-Deep Fool (Moosavi-Dezfooli et al. 2016), PGD-Projected Gradient Descent (Madry et al. 2017), NEW-Newton Fool (Jang et al. 2017), ELA-Elastic Net Attack (Chen et al. 2018). We select the black-box attacks: HOP-HopSkip Jump Attack (Chen et al. 2020), ZOO-Zeroth-Order Optimization Attack (Chen et al. 2017), BOUND-Boundary Attack (Brendel et al. 2017). For each attack, we apply many different parameters configurations to have attack samples with different perturbations: for each attack, images range from very little perturbation to evident perturbation.

We generate adversarial images from the first 100 images in the CIFAR test set. We exclude the images that are originally misclassified by the ConvNet12 model because in this case, the effect of the adversarial perturbation is not relevant. Further, we only include attacked images that effectively deceive the image classifier because we are interested in the detection of adversarial images that affect the classification. Consequently, we obtain a set of adversarial images that have 100% of success rate on the ConvNet12 model. In total we obtain 222 attack configurations and 18062 adversarial images.

### Research Questions

We study, through an extensive experimental campaign, which are the most suitable metrics to select the adversarial attacks when testing defenses. We search metrics that can

relate the attacks to the defense. We formulate the following questions:

*RQ1. Is a single L-norm sufficiently informative to use in the evaluation of adversarial defenses?* $L_0$, $L_1$, $L_2$, and $L_\infty$ norms are widely used: we investigate if these metrics allow measuring the capability of an adversarial image in evading defensive detectors.

*RQ2. Is it possible to identify a correlation between the output of an adversarial image detector and the value of a single L-norm?* In other words, we investigate if any L-norms can predict the output of the detector just by looking at the image perturbation.

*RQ3. Alternatively to L-norms (see RQ1), is there an image quality metric that allows to efficiently describe the strength of an attack?* In other words, we suggest the possibility of more informative alternatives with respect to L-norms.

*RQ4. Does combining L-norms with image quality metrics build a feature set that allows an accurate prediction of the output of a detector with respect to a given image?* In other words, can we increase the performance of the predictor built in *RQ2* using multiple metrics?

*RQ5 Is it meaningful to classify attacks based on the perturbation introduced in the image?* We investigate if different attacks generate similar perturbations that lead to similar outputs of the detector. If this is true, it would be possible to create groups of attacks, such that representatives from each group should be selected to properly cover the attack space.

To seek our answers, we need to compute metrics on adversarial images. We apply L-norms and image quality metrics to the adversarial images described previously. We use the implementation of the quality metrics provided by the *Sewar*[1] library, while L-norms implementations come from *sklearn*. After, each generated adversarial image is tested against two state-of-the-art detectors. We associate each image with the detection labels provided by the two detectors. Each label is 1 if the attacked image has been detected by the detector, and 0 on the contrary. Summarizing, we achieve a dataset that associates each adversarial image to two detection labels and the computed metrics values.

The two detectors are MagNet (Meng et al. 2017) and Squeezer (Xu et al. 2018) because i) they are very well-known, and ii) the source code is publicly available. MagNet (Meng et al. 2017) calculates the distance between the input x and a reformed input x'. If the image is genuine, the distance between x and x' is expected below a target threshold. If it is greater than a such threshold value, the input x is considered an attack. The threshold is selected during training, to target a specific false positive rate on the validation set. Squeezer (Xu et al. 2018) combines multiple feature squeezers. It compares the prediction of the classifier on the inputs with the predictions obtained using pre-processed inputs. The detector computes a score that is the maximum distance among these predictions. The final detection is done by selecting a pre-defined threshold for a such score, to distinguish between genuine and adversarial images. The training of the detector consists in selecting a threshold on the score.

### Execution Environment and Repositories

Experiments have been executed on a Dell Precision 5820 Tower with an Intel I9-9920X, GPU NVIDIA Quadro RTX6000 with 24GB VRAM, 128GB RAM, Ubuntu 18.04 with kernel 5.4.0, and runtime CUDA 11.0.

All data generated, the configuration parameters, and the code to reproduce our experiments is at [2].

### Experiments Execution and Results

We elaborate the answers to the 5 questions.

*Answer to RQ1.* We seek for adversarial images which show similar values of the L-norms but different responses from the detector. We found that, in some examples, the value a single norm is not enough to describe the effectiveness of the attack.

| At-tacks | L-norms | | | | Detectors Labels | |
|---|---|---|---|---|---|---|
| | L2 | L1 | L∞ | L0 | MagNet | Squeezer |
| DEEP | 10.717 | 477.524 | 0.682 | 3072 | 0 | 0 |
| BIM | 10.728 | 478.13 | 0.673 | 3072 | 1 | 1 |
| CW2 | 33.87 | 1793.76 | 0.867 | 3072 | 1 | 1 |
| ELA | 33.87 | 1795.76 | 0.864 | 3072 | 0 | 0 |

Table 1: Examples of adversarial images from the attack sets.

The attacks in the first two row from Table 1 have an $L_2$ norm close to 10.7; however, BIM attack is detected by MagNet and Sqeezer (detector labels on the right of the table are 1 in the second row), while DEEP is not detected (detector labels are 0 in the first row). The same behavior can be observed in the ELA and CW2 highlighted in gray. They have the exact same $L_2$ value, but an opposite decision from the detectors. Furthermore, in both attacks, the $L_1$, $L_0$, and $L_\infty$ values are very close to each other, if not equal. This makes it difficult to determine the effectiveness of an attack using a single norm. The values of the $L_1$ and $L_\infty$ norms are far from each other but, again, the detection label is the same.

*Answer to RQ2.* We research a statistical correlation between the results obtained by the detectors and the values of the selected metrics.

---

[1] Andrew Khalel, Sewar library, last modified on 12/07/2022, github.com/andrewekhalel/sewar

[2] Tommaso Puccetti, Experiments Repository, 2022, github.com/TommasoPuccetti/adversarial_perturbation

First, we compute the Pearson Correlation Coefficient between each metric and the labels of the detectors, but without relevant results. Then we use RF as a correlator: we train an RF classifier using part of the dataset created with the aim of predicting the response of the detectors. The idea is to predict the output of a detector to a specific image knowing only information about the perturbation described by the L-norms. We use 66% of the dataset as the training set, leaving the rest for the test set.

| Norm Metrics | Random Forest Accuracy | |
|---|---|---|
| | Magnet | Squeezer |
| $L_0$ | 0.695 | 0.733 |
| $L_1$ | 0.836 | 0.832 |
| $L_2$ | 0.833 | 0.823 |
| $L_\infty$ | 0.761 | 0.785 |

Table 2: Accuracy of the Random Forest calssifier trained using a single L-norm.

The results in Table 2 show that the L-norms have some ability to predict the response of the detectors, obtaining 0.836 accuracy in the best case. For example, training with $L_2$, the Random Forest can guess with an accuracy of 0.833 if an image is classified as an attack or normal image by MagNet. Similar capabilities are shown for Squeezer with an accuracy of 0.823. We obtained the worst results using $L_0$ with 0.695 and 0.733 of accuracy for MagNet and Squeezer respectively. The results are not surprising since $L_0$ measures the number of pixels to which a perturbation has been applied but it does not say anything about the perturbation intensity. The $L_\infty$ norm is more effective than the $L_0$. Also, in this case, the result is not surprising: the $L_\infty$ norm measures the maximum perturbation applied to each pixel but does not explain how the perturbation is distributed among the pixels. The $L_2$ can remain small when there are many small changes to many pixels, thus it is more informative about the overall perturbation. The $L_1$ distance has a capability comparable to the $L_2$. However, for all the L-norms considered, the correlation ability is not surprisingly high.

| Image Quality Metrics | Random Forest Accuracy | |
|---|---|---|
| | Magnet | Squeezer |
| PSNR-B | 0.852 | 0.882 |
| VIF | 0.834 | 0.878 |
| RASE | 0.827 | 0.877 |
| ERGAS | 0.823 | 0.87 |

Table 3: Accuracy of the Random Forrest classifier trained using image quality metrics.

*Answer to RQ3.* We repeat the same approach of RQ1 and RQ2 but with the image quality metrics. We identify contradictory examples like in RQ1 suggesting that, as for L-norms, solely computing quality metrics is not enough informative. Individually, the correlation is similar to the best L-norm. In fact, in Table 3 we report the image quality metrics that make the RF reach the top 4 accuracy scores. Table 3 shows that correlation with quality metrics is more evident than using the best L-norms $L_1$ and $L_2$. In particular, PSNR-B shows a significant increase in accuracy with respect to an RF using $L_1$.

*Answer to RQ4.* Since each norm gives us different information about the perturbation, we combine them to calculate more precise evaluations of the effectiveness of the attack. We repeat the experiment of RQ2 but including various norms.

First, we try all the possible permutations of the L-norms as features of the training set. The RF is then exercised against the test set to evaluate its accuracy in guessing the response of the detector. Results in Table 4 show that the more metrics we include in our training set, the more we can predict the output of the detectors on a given adversarial image, reaching a correlation accuracy above 0.91.

| Norm Metrics | Random Forest Accuracy | |
|---|---|---|
| | Magnet | Squeezer |
| L0, L1, Linf | 0.889 | 0.893 |
| L1, L2, Linf | 0.912 | 0.913 |
| L0, L2, Linf | 0.889 | 0.889 |
| L0, L1, L2, Linf | 0.912 | 0.915 |

Table 4: Accuracy of the Random Forest classifier trained combining multiple L-norms.

Then, we investigate image quality metrics. First, we exercise a feature selection method to understand which features are the most informative. The importance of the metrics as features change according to the detector, but MSE and PSNR-B are the most important for both detectors. We use the results of the feature selection method to select the metrics to combine. The combination we choose and their results in terms of RF accuracy are in Table 5.

The combination of the most important quality metrics for both detectors shows some improvement in accuracy (from 0.912 to 0.929 for Magnet, from 0.915 to 0.936 for Squeezer) with respect to the best results of Table 4. Combining all the quality metrics we get even better performances. We achieve an accuracy of 0.936 for Magnet and 0.949 for Squeezer. This suggests that the image quality metrics can be more informative on the attacks and their perturbation with respect to the L-norms.

| Detectors | Image Quality Metrics | Random Forest Accuracy |
|---|---|---|
| Magnet | MSE, SCC, VIF, PSNR-B | 0.929 |
| | ALL_QUALITY | 0.936 |
| Squeezer | MSE, ERGAS, SAM, PSNR-B | 0.936 |
| | ALL_QUALITY | 0.949 |

Table 5: Accuracy of the Random Forest using the best combination of four quality metrics and including all of them (ALL_QUALITY).

Last, we investigate L-norms and image quality metrics together. The good performances of the MSE and PSNR-B metrics are confirmed but, as in the previous case, the most relevant metrics for both detectors are different. In Table 6, we show the accuracy achieved by the RF classifier for the 4 most informative metrics with Magnet and Squeezer. Accuracy is, respectively, 0.931, and 0.936. The score for Magnet is obtained using quality metrics MSE and PSNR-B along with $L_2$ and $L_1$. This combination allows obtaining a slight increase in accuracy than those obtained using the 4 best quality metrics of Table 4 or the L-norms of Table 5. In the case of Squeezer, on the contrary, the most relevant features are the same quality metrics of Table 5 (MSE, ERGAS, SAM, PSNR-B). In this case, therefore, there is no benefit from enriching the feature set with the L-norms. The difference between feature importance in Magnet and Squeezer could be related to the different functional behavior of the two detectors.

| Detectors | Image Quality Metrics + L-Norms | Random Forest Accuracy |
|---|---|---|
| Magnet | MSE, PSNR-B, $L_2$, $L_1$ | 0.931 |
| | NORM + QUALITY | 0.937 |
| Squeezer | MSE, ERGAS, SAM, PSNR-B | 0.936 |
| | NORMS + QUALITY | 0.947 |

Table 6: Accuracy of the Random Forest using the best combination of metrics and combining all the metrics (either L-norms or quality metrics)

Last, we combine L-norms and all 8 image quality metrics achieving an accuracy of 0.937 for Magnet and 0.947 for Squeezer (see ALL_QUALITY in Table 5), which is comparable with the results obtained using only quality metrics. Thus, based on the correlation between the detection label and image perturbation, the image quality metrics can describe the strength of the attack even better than L-norms.

*Answer to RQ5.* We investigate if the response of the detector on unknown attacks can be predicted using only the RF trained on known attacks, i.e., attacks included in the training set. If so, it would mean that different attacks introduce alterations described in the same way by image quality metrics, which are detected similarly.

To achieve this goal, we rely on a leave-one-out approach where we create different test sets consisting of a single type of attack and, at the same time, exclude such attack from the training set. This process is repeated for each of the 10 attack types. Table 7 shows the accuracy of the RF against unknown attacks.

Results are heterogeneous. Table 7 reports the attacks that resulted in a high accuracy in gray: in this case, the accuracy is comparable to the values in the previous tables. The three black box attacks (BOUND, HOP, ZOO) plus three white box (ELA, PGD, BIM) have a high correlation.

For example, when BIM is the unknown attack, the accuracy is 1 for both detectors: our RF can perfectly predict the behavior of the detectors on BIM attacks just by measuring the perturbation introduced on known attacks. This lets us suppose that the alteration of the BIM images is similar to alterations in the training set.

This suggests that we are able to identify groups of attacks with similar perturbations and similar strength. In practice, this can result useful to evaluate the originality and uniqueness of attacks (not in terms of their mathematical formulation, but in the practical modifications to the images). At the same time, this way of reasoning could help reducing the attack types to be tested against a defense: if attacks can be grouped based on image perturbation and detector outputs, only one representative from each group could be selected.

| Unknown Attack | Magnet | Squeezer |
|---|---|---|
| BIM | 1 | 1 |
| BOUND | 0.874 | 0.95 |
| CW2 | 0.404 | 0.319 |
| DEEP | 0.662 | 0.708 |
| ELA | 0.932 | 0.959 |
| FGM | 0.64 | 0.682 |
| HOP | 0.942 | 0.924 |
| NEW | 0.759 | 0.746 |
| PGD | 0.948 | 0.756 |
| ZOO | 0.78 | 0.79 |

Table 7: Accuracy of Random Forest on unknown attacks.

The accuracy is much lower when the unknown attacks are FGM, DEEP, and CW2. The worst results are obtained using CW2 (accuracy is 0.40 on Magnet and 0.31 on Squeezer). This suggests that the perturbations introduced by FGM, DEEP, and CW2 are different from any other attack. Clearly, this consideration is valid under the assumption that the 12 metrics are descriptive enough to capture the characteristics of the introduced perturbation. To support our assumption, we remark that we used the most relevant image quality metrics from the state-of-the-art and the most used L-norms.

Summarizing, in many cases, the information included in the image provides an indication of attack detectability. This suggests that an RF classifier, trained on multiple attacks as we did, could be used to verify the strength of an attack against a target defense. In addition, it could be used to classify attacks based on the image perturbation (as measured by the 12 metrics) and the detectability, to assure that attacks from multiple classes are selected.

## Acknowledgments

This work has been partially supported by the NextGenerationEU programme, Italian DM737 – CUP B15F21005410003.

# Takeovers and Future Works

We investigated the information contained in L-norms and image quality metrics and their possible use to support the evaluation of adversarial attack defenses.

We observe that, under several circumstances, image quality metrics are more informative than L-norms. Further, we observe a correlation between the detector performance and the values of the identified metrics, to the extent that i) they can predict the response of a detector, and ii) they can classify attacks based on the perturbations introduced and their strength (ability to evade defenses).

In fact, we reached almost 95% of correlation accuracy between the values of the metrics and the detector outputs.

As future works, we are extending our approach to craft a methodology that allows grouping attacks based on the measured perturbation by L-norms, quality metrics, and defense output, such that representative attacks selected from the different groups can justifiably cover the attack space at the state of the art. In fact, attacks with significantly different values have been proven to be poorly correlated i.e., leading to very different detection outputs. To such an extent, the preliminary results adopted in this work will be extended by considering additional detectors, classifiers, and datasets.

Also, the RF classifier could be a handy tool to estimate the strength and novelty (in terms of introduced perturbation) of an attack toward a target model. In fact, adversarial images could be fed to the trained RF classifier without the need to install defenses (Magnet and Squeezer in our case). The advantage is that the configuration and installation of a defense is a difficult task: it must be adapted to the target architecture, it requires the installation of many packages, and, often, a good deal of critical thinking to optimize the choice of parameters.

Additionally, in this work, we did not consider the success rate of an adversarial attack in evading a classifier. This way we will investigate if the identified metrics can also predict the success of an attack on the target classifier in addition to the detection capability of a defense.